\documentclass{article}

\usepackage[nonatbib, final]{neurips_2021}
\usepackage[sort, compress, square, numbers]{natbib}
\usepackage[utf8]{inputenc} % allow utf-8 input
\usepackage[T1]{fontenc}    % use 8-bit T1 fonts
\usepackage{hyperref}       % hyperlinks
\usepackage{url}            % simple URL typesetting
\usepackage{booktabs}       % professional-quality tables
\usepackage{amsfonts}       % blackboard math symbols
\usepackage{nicefrac}       % compact symbols for 1/2, etc.
\usepackage{microtype}      % microtypography
\usepackage{xcolor}         % colors
\usepackage{amsmath}
\usepackage{amssymb}
\usepackage{url}
\usepackage{makecell}
\usepackage{multirow}
\usepackage{booktabs}
\usepackage{array}

\usepackage[utf8]{inputenc}

\usepackage{microtype}
\usepackage{kotex}

\newcommand{\thickhline}{%
    \noalign {\ifnum 0=`}\fi \hrule height 1pt
    \futurelet \reserved@a \@xhline
}

\usepackage{comment}
\usepackage{todonotes}
\usepackage{xcolor}
\title{How should human translation coexist with NMT? Efficient tool for building high quality parallel corpus}

\author{%
\parbox{\linewidth}{\centering
  Chanjun Park$^{1}$, Seolhwa Lee$^{2}$, Hyeonseok Moon$^{1}$ \\
  Sugyeong Eo$^{1}$, Jaehyung Seo$^{1}$, Heuiseok Lim$^{{1}\dagger}$} \\ \\
  $^{1}$ Korea University, \{bcj1210, glee889, djtnrud, seojae777, limhseok\}@korea.ac.kr \\
  $^{2}$ University of Copenhagen, sele@di.ku.dk \\
  
}

\begin{document}

\maketitle
% \footnotetext{\footnote[2]{} Corresponding author}
\begin{abstract}
This paper proposes a tool for efficiently constructing high-quality parallel corpora with minimizing human labor and making this tool publicly available. Our proposed construction process is based on neural machine translation (NMT) to allow for it to not only coexist with human translation, but also improve its efficiency by combining data quality control with human translation in a data-centric approach.
\end{abstract}

\section{Introduction}
Building a \textit{high-quality} parallel corpus, which has its target sentence precisely translates the source sentence, vice versa, is a common important issue in the entire field of machine translation. Unfortunately, obtaining a high-quality parallel corpus is difficult for many reasons, including problems of copyright acquisition, the difficulty of finding the proper alignment, and high monetary and temporal costs of building the corpus~\citep{koehn2020findings}. Human translation is fundamentally the most trusted approach for improving data quality, and it can construct a high-quality parallel corpus~\citep{hutchins2001machine,rojo2018aspects}. However, even this approach is limited, as it requires a tremendous amount of money and time for humans to manually construct the entire corpus. To alleviate this limitation, we present a novel tool for constructing high-quality parallel corpora using only simple mono corpus.  

In detail, we divided the generic process of constructing a high-quality parallel corpus into two components. First, a data advancing automation approach is employed to advance the source ({\em{i.e.,}} the initial mono corpus) language using corpus filtering~\citep{herold2021data} and Grammar Error Correction (GEC)~\citep{wang2020comprehensive}. Subsequently, the quality of the translated target is also improved by both the advanced mono corpus and the process of Automatic Post Editing (APE)~\citep{chatterjee2019findings}. Second, we utilize both the predicting automation approach ({\em{i.e.,}} predictions of data quality) and human translation to minimize the human labor required to complete the task. The predicting automation approach employs Quality Estimation~\citep{fonseca2019findings} to predict the sentence quality of the parallel corpus, and its labels are used to measure the human labor cost.

Thus, the limitations associated with generating a parallel corpus can be alleviated if the computer automatically determines the quality of the corpus according to a specified level of threshold control. That is, human labor is unnecessary when the threshold is exceeded, although if threshold is not exceeded then it requires refinement, which is the verification and post-processing of the corpus conducted by humans. Overall, our approach contributes to the human translation market and to the automated machine translation field by improving efficiency and minimizing cost.

\begin{figure}[tbh]
    \centering
    \includegraphics[width=0.80\textwidth]{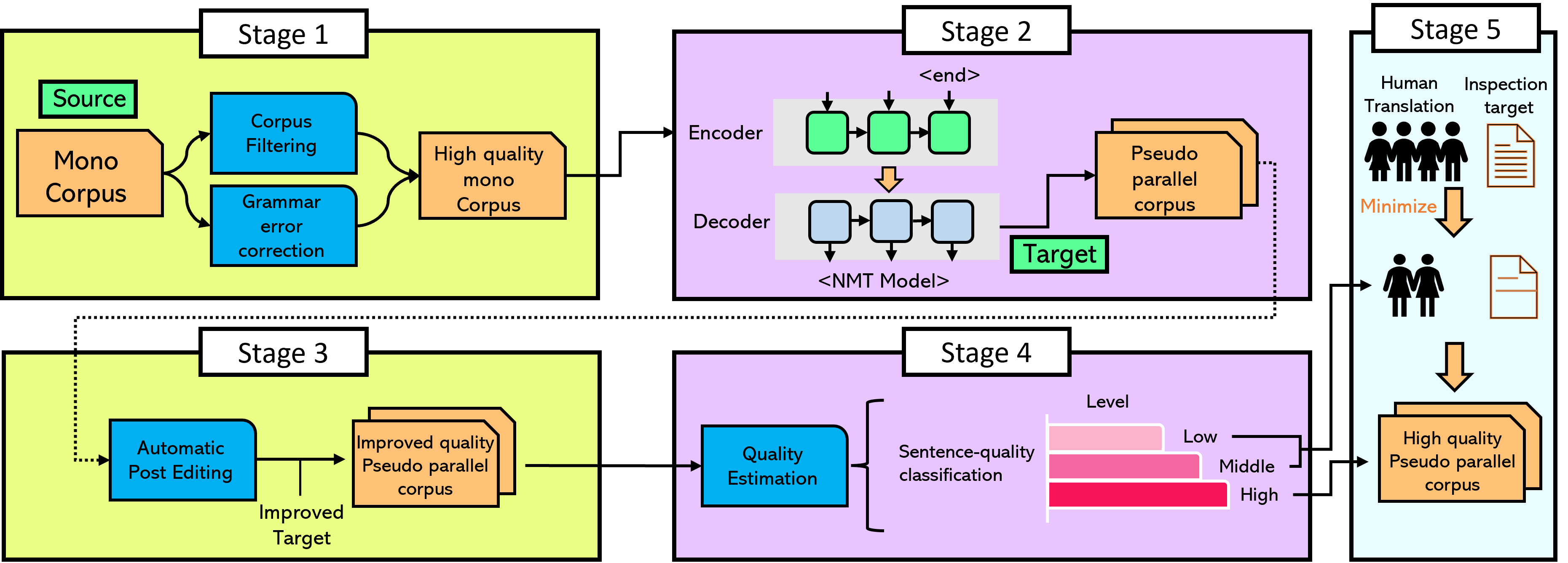}
    \caption{Overall process of building the high-quality parallel corpus based on our proposed tool.}
    \label{fig:overall}
\end{figure}

\section{Data Construction Process and Tool}
\paragraph{Process} 
Data construction processes that build high quality parallel corpus solely by mono corpus are described as Figure~\ref{fig:overall}.
 
For \textbf{Stage 1}, corpus filtering \citep{herold2021data} and grammar error correction \citep{wang2020comprehensive} are conducted to ensure the quality of the mono corpus. Through Stage 1, quality improvements on the source data is done automatically. In \textbf{Stage 2}, the refined mono corpus is translated ({\em{i.e.,}} the target) by the NMT model. Any well-performing NMT model can be utilized, either an in-house NMT model or commercialized translation system. Through this process, the primary pseudo-parallel corpus is constructed. In \textbf{Stage 3}, the Automatic Post Editing (APE) system corrects errors that exist in the primary pseudo-parallel corpus \citep{do2021review}. This phase contributes to the enhancement of the parallel corpus quality, particularly for the translation results from the target side. \textbf{Stage 4} proceeds with a quality prediction of the machine translation for the parallel corpus through the Quality Estimation (QE) model \citep{wang2020hw}. Automatic labeling of sentence quality (i.e, level) is performed based on Pearson's correlation, Mean Average Error (MAE), and Root Mean Squared Error (RMSE), which are sentence-level performance evaluation measures. The average value of the three metrics is selected as the final quality value. \textbf{Stage 5} determines the level of quality of the given inspection target using the corresponding score obtained from Stage 4. The continuously calculated score is quantized into three levels (High, Middle, and Low), and each sentence pair is classified based on these levels. We implemented a heuristic logic-based decision criteria that grouped sentences into those with scores over 20\% (High), under 20\% (Low), and between these values (Middle). The high-scoring level is regarded as a high-quality parallel corpus and can be used immediately, without modification. Sentence pairs included in the middle and low levels are assigned to the human translation supervisor to enhance their quality. The price for the editing labor can be estimated by the quality level that has been determined by the QE model. Therefore, the user decides whether to use each sentence as corpus data or have the translation supervised by a human agent at an agreed upon price.

The advantage of this process is that it enables the quantitative estimation of the data quality before translation, thereby the reducing the cost of human translation supervision, as the easier sentences are already translated by the machine translation system. For high level sentences, only minor supervision at most is required, whereas for low level sentences, relatively more intensive and in-depth editing will be required. Overall, this strategy can shorten the time required for editing and improve the efficiency of the supervision work.

\paragraph{Tool}
We implemented and distributed this tool in the form of a web application. The webserver was developed based on Flask. The corpus cleaning was implemented based on \citet{parkbts} filtering process and \citet{park2020comparison} GEC model. The Google Translator API was used for the NMT model. We developed and reimplemented the APE system based on the model in \citet{yang2020hw} and the QE system released by Transquest \citep{ranasinghe2020transquest} in the form of a Rest API and combined it with the tool. We released this tool to be publicly available\footnote{\url{http://nlplab.iptime.org:9090/}}.

\section{Conclusion} 
This paper proposed a data construction method that can work alongside the human translation market of machine translation. 
For future work, we plan to enhance the tools performance by improving the modular performance of each stage.

\section*{Acknowledgment}
This research was supported by the MSIT(Ministry of Science and ICT), Korea, under the ITRC(Information Technology Research Center) support program(IITP-2018-0-01405) supervised by the IITP(Institute for Information \& Communications Technology Planning \& Evaluation) and IITP grant funded by the Korea government(MSIT) (No. 2020-0-00368, A Neural-Symbolic Model for Knowledge Acquisition and Inference Techniques) and Basic Science Research Program through the National Research Foundation of Korea(NRF) funded by the Ministry of Education(NRF-2021R1A6A1A03045425).Thanks to Seungjun Lee for helping us build the Tool. Heuiseok Lim$^\dagger$ is a corresponding author.

\bibliography{ref}
\bibliographystyle{acl_natbib}

\end{document}